\documentclass[10pt,twocolumn,letterpaper]{article}

\usepackage{iccv}              % To produce the CAMERA-READY version
\usepackage[accsupp]{axessibility}
\usepackage{colortbl}
\usepackage[dvipsnames]{xcolor}

\usepackage{graphicx}
\usepackage{amsmath}
\usepackage{amssymb}
\usepackage{booktabs}

\usepackage{times}
\usepackage{epsfig}

\usepackage{verbatim}
\usepackage{amsmath,bm}
\usepackage{algorithm}
\usepackage{algorithmicx}
\usepackage{algpseudocode}
\usepackage{float} 

\usepackage{color}

\usepackage{ctable}
\usepackage{makecell}
\usepackage{tabularx}
\usepackage{multirow}
\usepackage{multicol}
\usepackage{arydshln}
\usepackage{afterpage}

\usepackage{wrapfig}
\usepackage{adjustbox}

\usepackage{color,xcolor}
\usepackage{colortbl}
\usepackage{enumitem}

\usepackage{pifont}

\definecolor{iccvblue}{rgb}{0.21,0.49,0.74}
\usepackage[pagebackref,breaklinks,colorlinks,allcolors=iccvblue]{hyperref}

\def\paperID{2524} % *** Enter the Paper ID here
\def\confName{ICCV}
\def\confYear{2025}

\title{ScoreHOI: Physically Plausible Reconstruction of \\ Human-Object Interaction via Score-Guided Diffusion}

\def\spaces{~~~~~~}
\author{Ao Li\textsuperscript{1,2}\spaces{}Jinpeng Liu\textsuperscript{1,2}\spaces{}Yixuan Zhu\textsuperscript{2}\spaces{}Yansong Tang\textsuperscript{1,2}\thanks{Corresponding Author}\\\\
\textsuperscript{1}Tsinghua Shenzhen International Graduate School\spaces{}
\textsuperscript{2}Tsinghua University
}

\begin{document}

\maketitle
% Remove page # from the first page of camera-ready.
% \ificcvfinal\thispagestyle{empty}\fi

%%%%%%%%% ABSTRACT
\begin{abstract}
Joint reconstruction of human-object interaction marks a significant milestone in comprehending the intricate interrelations between humans and their surrounding environment. Nevertheless, previous optimization methods often struggle to achieve physically plausible reconstruction results due to the lack of prior knowledge about human-object interactions. In this paper, we introduce ScoreHOI, an effective diffusion-based optimizer that introduces diffusion priors for the precise recovery of human-object interactions. By harnessing the controllability within score-guided sampling, the diffusion model can reconstruct a conditional distribution of human and object pose given the image observation and object feature. During inference, the ScoreHOI effectively improves the reconstruction results by guiding the denoising process with specific physical constraints. Furthermore, we propose a contact-driven iterative refinement approach to enhance the contact plausibility and improve the reconstruction accuracy. Extensive evaluations on standard benchmarks demonstrate ScoreHOI’s superior performance over state-of-the-art methods, highlighting its ability to achieve a precise and robust improvement in joint human-object interaction reconstruction. Code is available at \url{https://github.com/RammusLeo/ScoreHOI.git}

% The essence of this task is to harness prior interaction knowledge and physical constraints to enable precise reconstruction of both human and object poses. 
\end{abstract}  

%%%%%%%%% BODY TEXT
\section{Introduction}

The task of joint 3D human-object interaction reconstruction entails the recovery of both human body mesh and the pose of the interactive object from images or videos. In recent decades, human mesh recovery (HMR)~\cite{hmr, zhu2024dpmesh, hmr2, prohmr, spin} has garnered significant attention within the research community, owing to its extensive range of applications. Concurrently, a growing body of research has underscored the importance of reconstructing the interaction between humans and objects, rather than merely focusing on the reconstruction of the human. This approach holds substantial potential for various applications, such as data collection for robotics, VR/AR and game development.

\begin{figure}[t]
    \centering
    \includegraphics[width=0.99\linewidth]{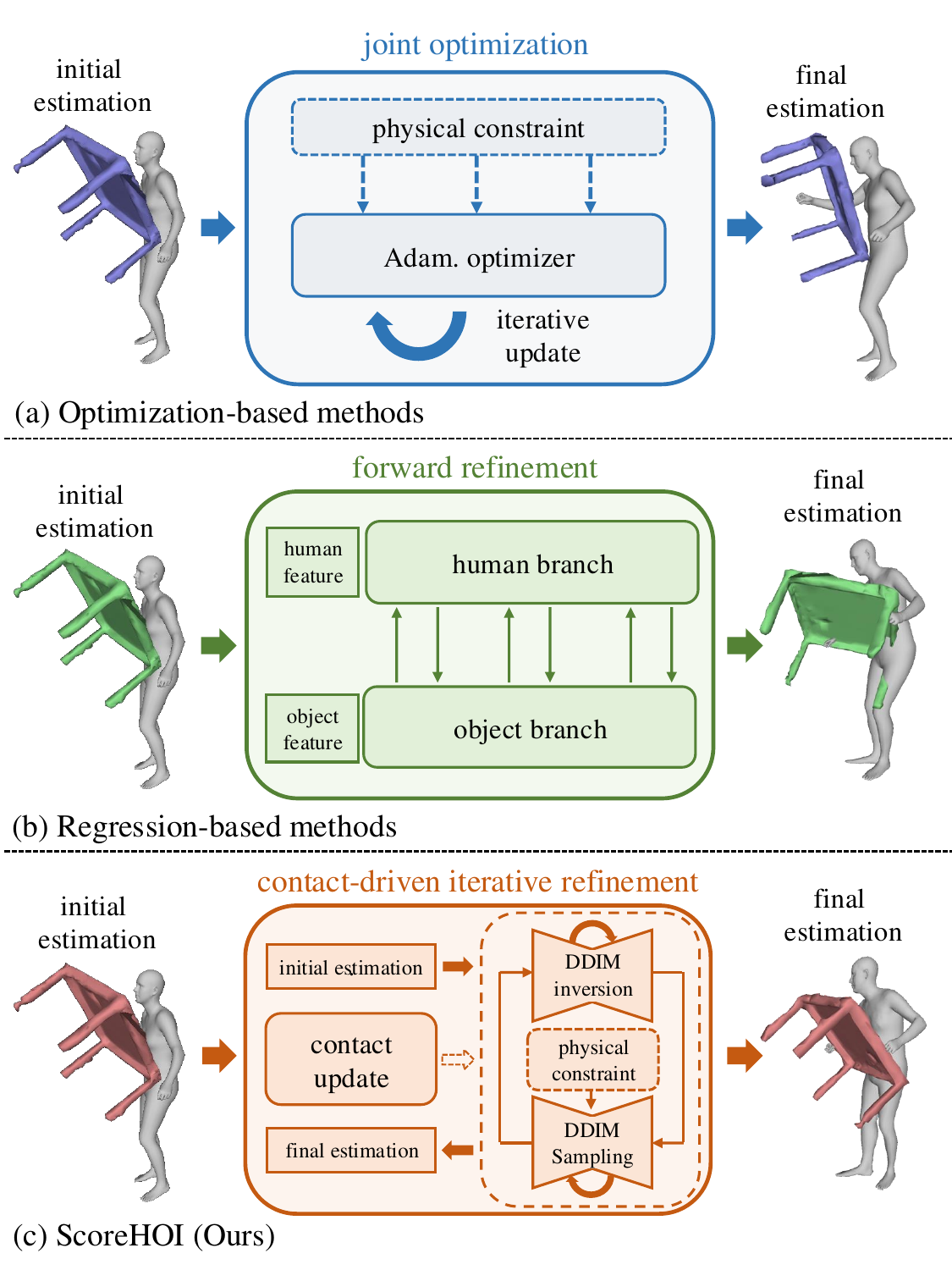}
    \vspace{-5pt}
    \caption{\textbf{Comparison of current methods and the proposed ScoreHOI.} (a) Optimization-based methods iteratively refine predicted outcomes with the physical objectives. (b) Regression-based methods update predictions via a dual-branch forward process. (c) Our ScoreHOI integrates a score-based denoising module that incorporates physical constraints during the sampling process.}
    \label{fig:teaser}
    \vspace{-15pt}
\end{figure}

Despite advancements, the direct reconstruction of human and object meshes from monocular images~\cite{zhang2020perceiving} remains a formidable challenge. This difficulty arises from the omission of numerous interaction patterns, such as grasping, lifting, or sitting, between humans and objects, which constitute valuable prior knowledge for the recovery task. To address this, existing methodologies have incorporated refinement modules that aim at transforming coarse estimations into more accurate outcomes. As illustrated in Figure~\ref{fig:teaser}, the prevailing approaches can be broadly categorized into optimization-based and regression-based methods. Optimization-based techniques~\cite{xie2022chore, xie2023visibility, ProciGen} employ a joint optimizer like Adam~\cite{kingma2014adam} to iteratively refine results by incorporating physical objectives, including contact and collision constraints. However, an overemphasis on physical constraints while neglecting image-level features often results in significant reconstruction inaccuracies. Moreover, these methodologies are often inefficient and require considerable computational time. Conversely, regression-based approaches~\cite{nam2024joint} devise a contact-aware cross-attention mechanism to reestablish the human-object relationship in a forward process that integrates image features. Nevertheless, the single-step forward refinement of parameters tends to lack robustness, particularly in scenarios like severe occlusion or depth ambiguity. Recently, diffusion models~\cite{ho2020denoising, song2020score, song2020denoising} have demonstrated their capability to recover data distributions from Gaussian noise. These models infer the implicit prior of the underlying data distribution $x$ by aligning the gradient of the log density $\nabla_x \log p(x)$~\cite{song2020score}, which is also referred to as the score of the distribution. Leveraging Bayes’ Theorem, the denoising process can be enhanced with an additional realistic guidance term, thereby steering the optimization process closer to the observed data. Owing to their rich prior knowledge and guided sampling proficiency, diffusion models have been employed to enhance the estimation quality in HMR tasks~\cite{zhu2024dpmesh, stathopoulos2024score, hmdiff2023distribution, lu2024dposerdiffusionmodelrobust}. More recently, methods like~\cite{xie2024template} endeavor to incorporate diffusion models into the human-object reconstruction. However, these approaches primarily seek to leverage diffusion techniques for the generation of point clouds, while we focus on exploiting the potential of diffusion priors for refining the human-object interaction.

To overcome the aforementioned limitations of prior approaches, we propose ScoreHOI, an effective framework that utilizes a score-based diffusion model to refine the coarse estimation results, as shown at the lower section of Figure~\ref{fig:teaser}. Our design contemplates two main challenges: (1) the integration of human-object interaction prior knowledge into the optimization process, and (2) the supervision of the sampling procedure with plausible physical constraints. 
To tackle these issues, we propose a contact-driven iterative refinement approach that leverages diffusion generative models as a robust optimizer endowed with extensive prior knowledge. In the inference phase, we first invert an initial regression estimation into a noisy latent representation via DDIM~\cite{song2020denoising} inversion. Subsequently, we denoise this latent by DDIM sampling, augmented with interaction guidance such as contact and collision constraints. To enhance the controllability, we introduce the object geometry and image feature as the conditions. Furthermore, we iteratively refine the contact masks to improve the physical plausibility. 

We conduct extensive experiments on standard benchmarks, including BEHAVE~\cite{behave} and InterCap~\cite{intercap} based on various initial regression results. Our ScoreHOI performs strong robustness and effectiveness, surpassing previous state-of-the-art methods and demonstrating significantly improved accuracy. Notably, we attain a 9\% improvement in the contact F-Score on the BEHAVE benchmark, thereby substantiating the optimization capability of our approach. In addition, we conduct comprehensive ablation studies to underscore the effectiveness and efficiency of our pipeline and the refinements we have designed.

%To tackle these issues, we harness the potential of diffusion models across both the training and inference phases. On one hand, we start with pre-training our diffusion model by large-scale human-object interaction datasets without image hints. Subsequently, we employ an OB-Adapter to integrate observation information into the denoising module. On the other hand, during inference, we first invert an initial regression estimation into a noisy latent representation via DDIM~\cite{song2020denoising} inversion. We then denoise this latent representation using DDIM sampling, augmented with interaction guidance such as contact and collision constraints.
\section{Related Work}

\noindent \textbf{Human-Object Interaction Reconstruction.} 
Modeling 3D interactions between humans and objects is a significant challenge. While recent studies have demonstrated strong performance in modeling human motion using high dimensional data ~\cite{contactgrasp, GRAB, promotion, omnicontrol, motionlcm, andriluka2014mpii,contactpose} and images~\cite{ganhand, use, learning, grasping}. However, these methods are limited to hand-object interactions and do not extend to full-body interactions, which are even more complex. Some approaches, such as PROX~\cite{resolving}, can fit a 3D human model to meet scene constraints~\cite{populating, pigraphs, place, couch}, capture interactions from multiple views~\cite{sun2021neural, jiang2022neuralhofusion, yi2022human}, or reconstruct 3D scenes based on human-scene interactions~\cite{yi2022human}. Recently, research has also expanded to include human-human interactions~\cite{three} and self-contacts~\cite{self, AAAI_learning}. Despite these advancements, existing methods struggle to jointly reconstruct human-object contacts from single images. A few studies have attempted to reason about 3D contacts from images~\cite{holistic++, cao2020long, huang2022capturing} and videos~\cite{estimating, imapper, contact, zhang2025flashvstream, wang2024hierarchical}, but they do not achieve the 3D reconstruction of full body and objects from a single RGB image. Weng et al.~\cite{holistic} predict the 3D human and scene layout separately, using scene constraints to refine human reconstruction. They apply predefined contact weights on SMPL-X~\cite{smplx} vertices, which can lead to inaccuracies. PHOSA~\cite{phosa} fits SMPL and object meshes separately and relies on predefined contact pairs to reason about interactions, but these heuristics are not scalable and often lack accuracy. Our work injects diverse physical constraints through contact-driven iterative refinement. It leads to more reasonable positions of objects and human mesh.

\begin{figure*}
    \centering
    \includegraphics[width=0.99\linewidth]{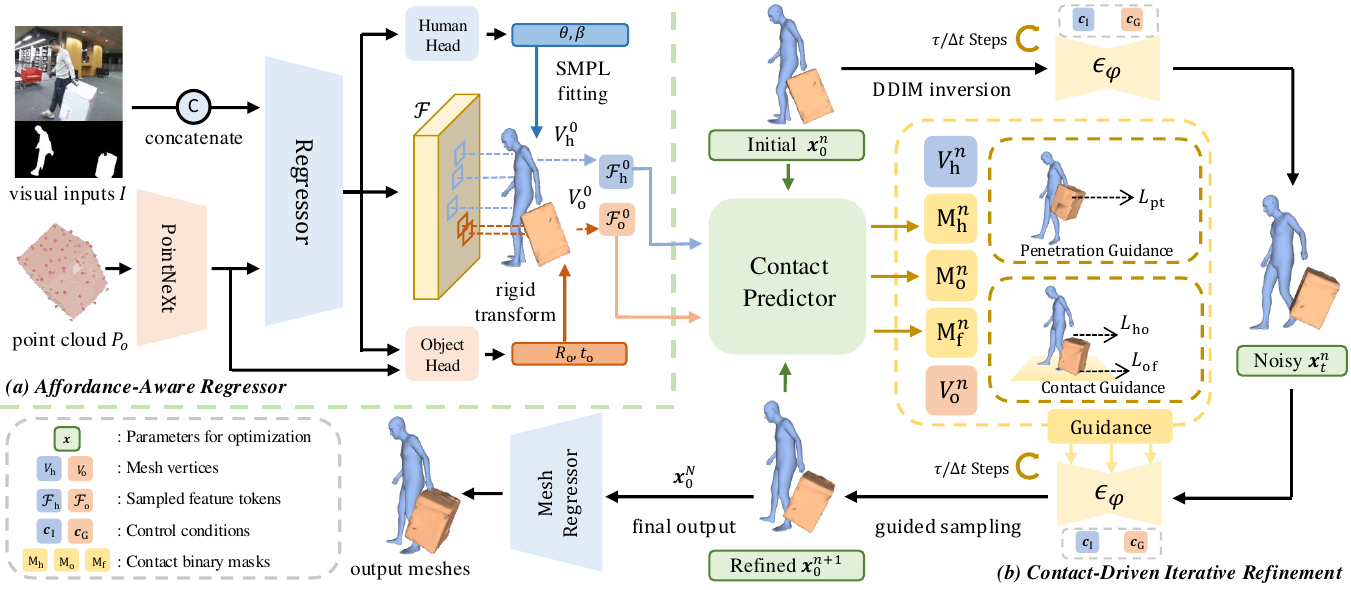}
    \vspace{-10pt}
    \caption{\textbf{The inference procedure of ScoreHOI.} (a) Given the input image $I$ the human and object segmented silhouette $S_{\rm h}, S_{\rm o}$ and the object template $P_{\rm o}$, we initially extract the image feature $\mathcal{F}$ and estimate the SMPL and object parameters $\theta$, $\beta$, $R_{\rm o}$ and $t_{\rm o}$. (b) Employing a contact-driven iterative refinement strategy, we refine these parameters $\bm{x}$ through the execution of a DDIM inversion and guided sampling loop. During the sampling process, physical constraints such as penetration $L_{\rm pt}$ and contact $L_{\rm ho}, L_{\rm of}$ are actively supervised. Following each optimization iteration, the contact masks $\{\mathbf{M}_{i}\}_{i \in \{ {\rm h}, {\rm o} ,{\rm f}\}}$ are updated to enhance the precision of the guidance.}
    \label{fig:pipeline}
    \vspace{-15pt}
\end{figure*}

\noindent \textbf{Score-Based Diffusion Models.}
Diffusion models~\cite{ho2020denoising, song2020score} have emerged as powerful tools for representing complex probability distributions, demonstrating exceptional performance in text-to-image generation~\cite{ho2020denoising,rombach2022high}. 
In addition to traditional denoising techniques, \cite{song2020score} introduces the concept of a score to characterize the time-dependent gradient field of the perturbed data distribution. The application of score-based sampling enables the denoising process to be directed by diverse forms of guidance, thereby fostering the development of methods applicable to a wide range of tasks, including super-resolution \cite{song2023pseudoinverse, zhu2024flowie, zhuinstarevive} and image inpainting \cite{improving, chung2022diffusion}. Recently, methods like~\cite{stathopoulos2024score} have explored the potential of the score-based diffusion models for HMR tasks by refining the outcomes with observations. However, the score-guided optimization in the context of 3D human-object reconstruction remains largely unexplored.
Our work aims to bridge this gap by integrating diffusion priors into this domain, which can potentially enhance the accuracy of reconstruction and the physical plausibility of interaction. 
% Notable applications include image inpainting~\cite{improving, chung2022diffusion}, super-resolution~\cite{song2023pseudoinverse}, and colorization~\cite{improving}, where these models have significantly improved the quality and realism of the generated outputs. 
% Our work aims to bridge this gap by investigating the applicability of score-guided sampling in the reconstruction of 3D human-object interaction, thereby contributing to the advancement of both theoretical understanding and practical implementations in this area.

\section{Method}

In this section, we present ScoreHOI, an effective optimizer designed to leverage priors and physical constraints for joint human-object interaction reconstruction. We will start by reviewing the background of human body models and score-based diffusion models. Then we will firstly introduce the full inference pipeline from the affordance-aware regressor to the contact-driven iterative refinement, as shown in Figure~\ref{fig:pipeline}. Following this, we will demonstrate the architecture of the diffusion model and other details.

\subsection{Preliminaries}

\noindent \textbf{Human Body Model.} 
SMPL~\cite{loper2015smpl} is a sophisticated parametric model designed to represent the human body in three dimensions. Two sets of parameters characterize this model: pose parameters $\theta \in \mathbb{R}^{24\times3}$, which capture the orientation and configuration of the body, and shape parameters $\beta \in \mathbb{R}^{10}$, which define the physical characteristics of the body such as size and proportions. The model establishes a mapping function $\mathcal{M}(\theta, \beta)$ that translates these parameters into a detailed body mesh $\mathcal{M} \in \mathbb{R}^{N\times3}$, where $N = 6980$ indicates the total number of vertices that comprise the mesh. 
% This approach not only enhances the fidelity of the model but also facilitates various applications in fields such as computer graphics, animation, and biomechanics. 
The SMPL-H~\cite{loper2015smpl, smplx, mano} model is based on a parametric representation, which combines the SMPL~\cite{loper2015smpl} body model and MANO~\cite{mano} hands model. This allows for the generation of realistic human figures that can be used in various applications, including manipulating objects with hands and interacting with the scene.

\noindent \textbf{Score-Based Diffusion Models.}
Diffusion models, such as DDPM~\cite{ho2020denoising}, comprise two processes: the \textit{forward diffusion} and the \textit{inverse denoising}. In \textit{forward diffusion}, data $\bm{x}_0$ is transformed into noised data $\bm{x}_T$ over $t=T$ timesteps using Gaussian kernels with variance schedule $\{\zeta_t\}$, defined by $q(\bm{x}_t | \bm{x}_0) = \mathcal{N} (\sqrt{\alpha_t} \bm{x}_0, (1-\alpha_t) \mathbf{I})$, where $\alpha_t := \prod_{s=1}^{t} (1 - \zeta_s)$. For the \textit{inverse denoising} process, a denoising model $\epsilon_{\phi}$ is trained to predict noise by minimizing:
$
\mathcal{L}(\mathbf{\phi}) = \mathbb{E}_{\bm{x}_0,t,\epsilon} ||\epsilon_{\phi} (\bm{x}_t, t) -  \epsilon||^{2},
$
where $t$ is uniformly sampled from $\{1, .., T\}$, and $\epsilon$ is the noise added to $\bm{x}_0$. According to~\cite{song2020score}, the model's predicted noise at timestep $t$ is linked to the score of the model:
\begin{equation}
    \epsilon_{\phi} (\bm{x}_t, t) = - \sqrt{1 - \alpha_t}  \nabla_{\bm{x}_t} \log p (\bm{x}_t),
    \label{eq: score}
\end{equation}
where $\nabla_{\bm{x}_t} \log p (\bm{x}_t)$ refers to the score, which describes the time-dependent gradient field of the perturbed data distribution.
For accelerated sampling, DDIM~\cite{song2020denoising} can be used, which employs non-Markovian processes. Based on~\ref{eq: score}, the denoised result $\hat{\bm{x}_0}(\bm{x}_t)$ from $\bm{x}_t$ is calculated as:
\begin{align}
    \hat{\bm{x}_0}(\bm{x}_t) &= \frac{1}{\sqrt{\alpha_{t}}} (\bm{x}_t - \sqrt{1 - \alpha_{t}} \epsilon_{\phi} (\bm{x}_t, t)) \\ \nonumber
    & \simeq \frac{1}{\sqrt{\alpha_{t}}} (\bm{x}_t+ (1 - \alpha_{t}) \nabla_{\bm{x}_t} \log p(\bm{x}_t)).
\end{align}
This framework also supports conditional distributions.

% Score-Based Diffusion Models~\cite{song2020score} are a class of generative models that generate high-quality data samples by gradually adding noise to the data and learning to reverse this process through a score function, defined as $s(x)=\nabla_{x}log_{p(x)}$, where $p(x)$ is the data distribution. The process consists of a forward diffusion phase described by a stochastic differential equation (SDE): 
% \begin{equation}
%     dx_t = \frac{1}{2}\beta(t)\nabla_xlog_{p(x_t)}dt + \delta(t)dW_t,
% \end{equation}
% where $\beta(t)$ controls the noise schedule, $\theta(t)$ is the noise level, and $dW_t$ is a Wiener process. This transforms the data into a simple noise distribution (often Gaussian) as $t \xrightarrow{} T$. The reverse denoising phase is modeled as:
% \begin{equation}
%     dx_t = (\nabla_xlog_{p(x_t)} - \frac{1}{2}\beta(t))dt + \delta(t)dW_t,
% \end{equation}
% where the learned score function guides the generation of new samples from noise. The model is trained using score matching techniques to estimate the score function, making it effective for various applications such as image synthesis and conditional generation.

\begin{algorithm}[t]
    \caption{Contact-Driven Iterative Refinement}
    \label{algo: contact}
    \begin{algorithmic}[1]
    \State \textbf{Input:} Initial parameters $\bm{x}^0_0$, diffusion model $\epsilon_{\theta}$, image feature $\mathcal{F}$, time steps $t$, and condition $\bm{c}$
    \State \textbf{Output:} The optimized parameters $\bm{x}^N_0$ 
    \For{$n = 0$ to $N-1$} 
        \State $\mathcal{F}^{n}_{\rm h}, \mathcal{F}^{n}_{\rm o} \gets \textbf{sample}(\bm{x}^n_0, \mathcal{F})$
        \State $\{\mathbf{M}_{i}\}_{i \in \{ {\rm h}, {\rm o} ,{\rm f}\}} \gets \textbf{Contact}(\bm{x}^n_0, \mathcal{F}^n_{\rm h}, \mathcal{F}^n_{\rm o})$ 
        \State $\bm{x}^{n+1}_0 \gets \textbf{DDIM\_Loop}(\bm{x}^n_0, t, \bm{c}, \epsilon_{\theta}, \{\mathbf{M}_{i}\}_{i \in \{ {\rm h}, {\rm o} ,{\rm f}\} })$
    \EndFor
    \State \Return $\bm{x}^N_0$
    \end{algorithmic}
\end{algorithm}

\subsection{Affordance-Aware Regressor}

For the joint human-object interaction reconstruction task, our primary goal is to precisely estimate the human body-hand pose $\theta \in \mathbb{R} ^ {52 \times 6}$, human shape parameter $\beta \in \mathbb{R} ^ {10}$, along with the object rotation $R_{\rm o} \in \mathbb{R} ^ {6}$ and the object translation $t_{\rm o} \in \mathbb{R} ^ {3}$.
We adopt the 6D rotation format following~\cite{zhou2019continuity} for the representation of $\theta$ and $\bm{R}_{\rm o}$ to enhance the stability of predicted results. We use the cropped image $I_{rgb}\in \mathbb{R}^{H \times W \times 3}$ with its corresponding human and object segmentation parts $S_{\rm h} \in \mathbb{R}^{H \times W \times 3}$ and $S_{\rm o} \in \mathbb{R}^{H \times W \times 3}$ as the visual inputs. Following previous work~\cite{xie2022chore, nam2024joint}, a coarse object template $P_{\rm o} \in \mathbb{R}^{64 \times 3}$ is given to initialize the shape of the object. However, in general situations, the shape of the object is varied in numerous forms, while the object in our training data is limited. Previous work like~\cite{nam2024joint} embeds the class identity to inject the object category information, yet this method is inaccessible when the object shape is out of the training set.

Although the objects have different shapes, we have a common assumption that the objects in the same category have a similar appearance, \textit{e.g.} a table has a flat surface and usually has legs supported on the ground. Therefore, to enhance the generalization ability, we introduce the Affordance concept inspired by~\cite{peng2023hoi}. The affordance refers to the perceived or actual properties of an object that suggest how it can be used, which can be constructed with a pre-trained affordance-aware network. The pre-trained model generates abundant prior knowledge from large scale 3D object datasets that is valuable for the reconstruction. With the help of affordance awareness, we extract the image feature $\mathcal{F}$, then we apply two separate heads to initially estimate the human body-hand SMPL-H parameters $\theta^0$ and $\beta^0$, as well as the object rotation $R^0_{\rm o}$ and translation $t^0_{\rm o}$.

\subsection{Optimization with Physical Guidance}
\label{sec: opt}

After obtaining the initial parameters, we define the optimization objective as the concatenation of human body pose, human body shape, object rotation and object translation, represented as $\bm{x} = \{ \theta, \beta, R_{\rm o}, t_{\rm o} \} \in \mathbb{R}^{331}$.

We start from the initial estimation result $\bm{x}^{\rm init}$, using DDIM inversion process to obtain the noisy latent $\bm{x}_{\tau}$, where $\tau$ is the noise level:
\begin{equation}
    \bm{x}_{t+1} = \sqrt{\alpha_{t+1}}\hat{\bm{x}_{0}}(\bm{x}_{t}) + 
    \sqrt{1- \alpha_{t+1}} \epsilon_{\phi}(\bm{x}_{t}, t, \bm{c}),
    \label{eq: ddim_inversion}
\end{equation}
where $\bm{c}$ is the combination of image feature condition $\bm{c}_{\rm I}$ and geometry feature condition $\bm{c}_{\rm G}$ defined in Section~\ref{sec: arch}.

After we have obtained $\bm{x}_{\tau}$, $\bm{x}^{\rm init}$ can be retrieved via the forward DDIM sampling process. Nevertheless, the intention is to enhance $\bm{x}^{\rm init}$ by incorporating physical constraints. Instead of using original score $\nabla_{\bm{x}_t} \log p(\bm{x}_t|\bm{c})$, we apply the conditional score with physical objectives $\nabla_{\bm{x}_t} \log p(\bm{x}_t|\bm{c}, \mathcal{P})$. According to the Bayes rule, the score can be written as $\nabla_{\bm{x}_t} \log p(\bm{x}_t|\bm{c}, \mathcal{P}) = \nabla_{\bm{x}_t} \log p(\bm{x}_t|\bm{c}) + \nabla_{\bm{x}_t} \log p(\mathcal{P}|\bm{c}, \bm{x}_t)$, where $\nabla_{\bm{x}_t} \log p(\bm{x}_t|\bm{c})$ is the output of the diffusion model. However, directly computing $\nabla_{\bm{x}_t} \log p(\mathcal{P}|\bm{c}, \bm{x}_t)$ from $\bm{x}_t$ is difficult. Inspired by~\cite{stathopoulos2024score, chung2022diffusion, scenediffuser}, we can take an assumption that:
\begin{equation}
    \nabla_{\bm{x}_t} \log p(\mathcal{P}|\bm{c}, \bm{x}_t) \simeq  \nabla_{\bm{x}_t} \log p(\mathcal{P}|\bm{c}, \hat{\bm{x}_0}(\bm{x}_t))
    \label{eq: diffusion}
\end{equation}
where $\hat{\bm{x}_0}(\bm{x}_t)$ represents the denoised results from $\bm{x}_t$. Consequently, the predicted human mesh $\hat{V^{\rm h}_0}(V^{\rm h}_t)$ and object mesh $\hat{V^{\rm o}_0}(V^{\rm o}_t)$ can be derived from $\hat{\bm{x}_0}(\bm{x}_t)$ to elucidate the interaction relationship, which constitutes the fundamental elements of computing physical constraints.

Now we aim to define the physical objectives, consisting of the human-object contact, object-floor contact and 3D penetration following~\cite{xie2023visibility, scenediffuser, chois}:
\begin{equation}
    L_{\mathcal{P}} = \lambda_{\rm ho}L_{\rm  ho}+ \lambda_{\rm of}L_{\rm of}+\lambda_{\rm pt}L_{\rm pt},
    \label{eq: objectives}
\end{equation}
where the individual constraints are given as:
\begin{itemize}[leftmargin=0.2em]
    \setlength{\itemsep}{0.2pt}
    \setlength{\parsep}{0.2pt}
    \setlength{\parskip}{0.2pt}
    % \item \textbf{Human reprojection}: $L_{\rm rh} = \y_{\rm conf} || \Pi_{\rm K}(V_{\rm h}) + \mathbf{\gamma} - \y_{\rm kp}||_{2}^{2}$,.
    % \item \textbf{Object reprojection}: $L_{\rm ro} = \sum^{N_{\rm o}}_{i=1} (1-\mathcal{G}(\Pi_{\rm K}(v^{\rm o}_i), S^{\rm o}))$. We tend to align the predicted object mesh $V_{\rm o}$ with the object segmentation silhouette $S_{\rm o}$ by computing the intersection over union~(IOU) $\mathcal{G}$. 
    \item \textbf{Human-Object Contact}: $L_{\rm ho} = || (\mathbf{M}_{\rm h} + \mathbf{M}_{\rm o}) \odot | V_{\rm h} - V_{\rm o} |~||_{2}$. The Euclidean distance between human mesh $V_{\rm h}$ and object mesh $V_{\rm o}$ at the predicted human contact regions $\mathbf{M}_{\rm h}$ and object contact regions $\mathbf{M}_{\rm o}$ should be ideally zero.
    \item \textbf{Object-Floor Contact}: $L_{\rm of} = || \mathbf{M}_{\rm f} \odot | V_{\rm o} |~||_{1}$. The $\mathbf{M}_{\rm f}$ represents the set of object vertices in contact with the ground surface. For vertices within this set, the height should be zero, indicating direct contact with the floor.
    \item \textbf{Penetration Avoidance}: $L_{\rm pt} = - \mathbb{E} [|\Phi^-_0 (V_{\rm h})|]$. We introduce a penalty for penetration scenarios utilizing the sign distance function (SDF) $\Phi^-_0$ of objects. The value of $L_{\rm pt}$ increases in the event of the overlaps happen.
\end{itemize}

Finally, the modified noise prediction with physical constraints is written as:
\begin{equation}
    % \epsilon'_{\phi} = \epsilon_{\phi}(\bm{x}_t, t, \bm{c}) + \rho \sqrt{1-\alpha_t} L_{\mathcal{P}},
    \epsilon'_{\phi} = \epsilon_{\phi}(\bm{x}_t, t, \bm{c}) + \rho\sqrt{1 - \alpha_{t}}\nabla_{\bm{x}^n_t}L_{\mathcal{P}}
    \label{eq: final_noise}
\end{equation}
where $\rho$ is the weight to control the guidance scale.

\subsection{Contact-Driven Iterative Refinement}
\label{sec: contact}

During the denoising process procedure guided by physical guidance, we find that the precise prediction of contact regions is a critical factor. 
The prediction of these regions is derived from the grid-sampled features $\mathcal{F}_{\rm h}$ and $\mathcal{F}_{\rm o}$ by human and object pose parameters $\bm{x}$. 
However, the straightforward estimation of the contact mask via a single forward process is prone to deficiencies, especially under heavy occlusion.
To address this, we have implemented a contact-driven iterative refinement strategy to augment the accuracy of contact estimation during the inference phase, as shown in Algorithm~\ref{algo: contact}.
Commencing with an initial parameter $\bm{x}^n_0$ at $n$-th iteration, we firstly sample the $\mathcal{F}_{\rm h}$ and $\mathcal{F}_{\rm o}$ from image feature $\mathcal{F}$. 
Subsequently, the human-to-object contact mask $\mathbf{M}_{\rm h}$, object-to-human contact mask$\mathbf{M}_{\rm h}$ and object-to-floor contact mask $\mathbf{M}_{\rm f}$ will be update by a contact predictor.
Then as mentioned in Section~\ref{sec: opt}, we obtain noisy $\bm{x}^n_{\tau}$ via DDIM inversion and generate refined $\bm{x}^{n+1}_0$ through guided sampling, wherein the contact masks and the estimated human and object vertices are incorporated. 
Assuming that we optimize for $N$ steps, the final iteration yields $\bm{x}^{N}_0$. 
Following~\cite{nam2024joint}, we apply a dual-branch masked transformer~\cite{vaswani2017attention} to ultimately enhance the quality of mesh recovery.

\subsection{Diffusion Model Architecture}
\label{sec: arch}

\begin{figure}
    \centering
    \includegraphics[width=0.99\linewidth]{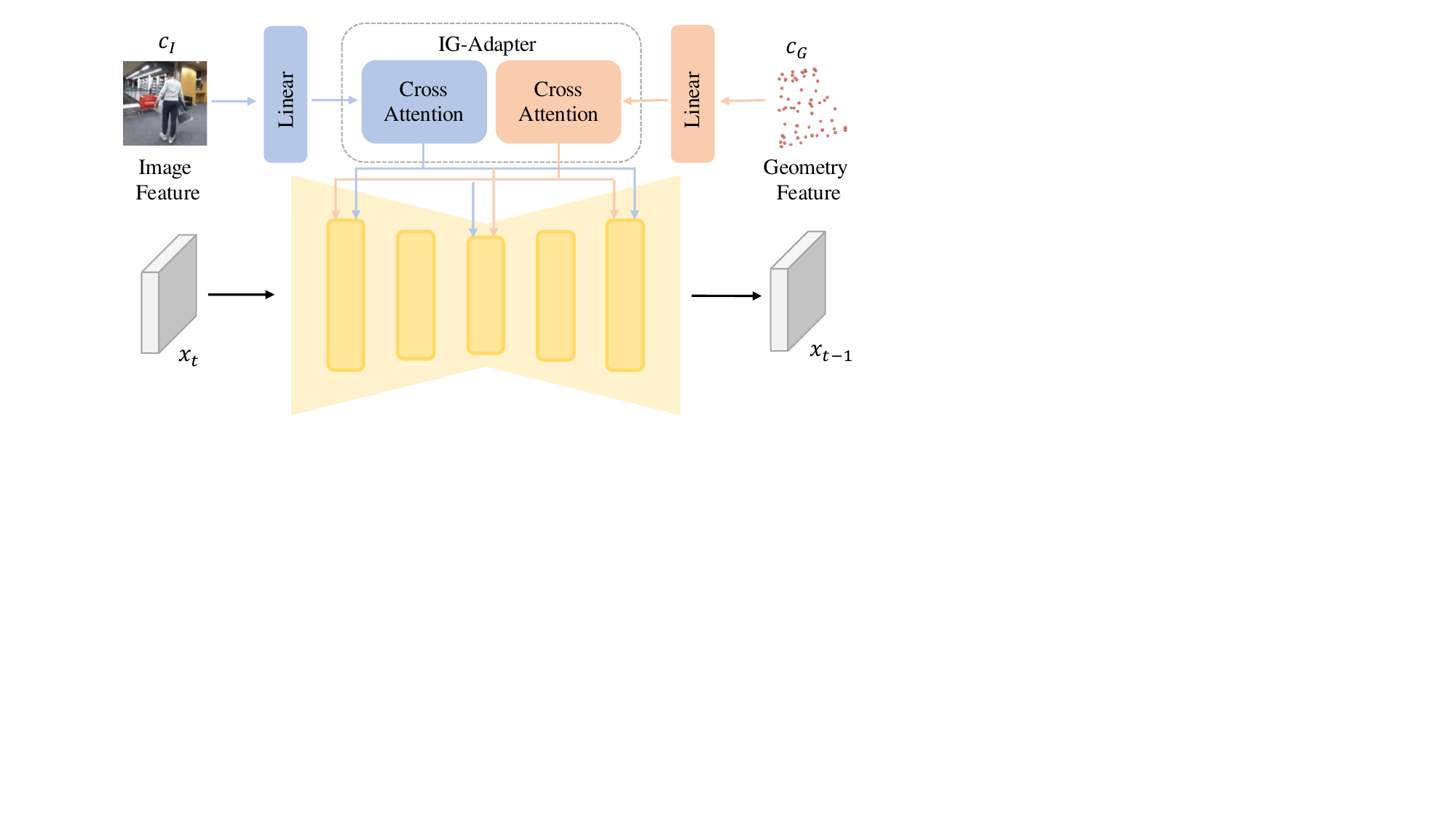}
    \vspace{-10pt}
    \caption{\textbf{The overview of IG-Adapter.} We introduce an IG-Adapter designed to integrate the image feature guidance $c_{\rm I}$ and the geometry feature guidance $c_{\rm G}$ into the diffusion model. The incorporation of observational and geometric awareness enhances the controllability of the model during the inference process.}
    \label{fig:fig2}
    \vspace{-10pt}
\end{figure}

The main diffusion model is built with 3 cross-attention layers. 
However, without any guidance, the generation results are not controllable, not to mention in reconstruction tasks.
To address this, we propose an IG-Adapter that can incorporate object geometry prior knowledge and visual observation as guidance.
We introduce two conditions including the image feature condition $\bm{c}_{\rm I}$ derived from the average pooling of $\mathcal{F}$ and the geometry feature condition $\bm{c}_{\rm o}$ extracted from pre-trained affordance-aware network. 
Inspired by~\cite{ipadapter}, we train an additional cross-attention block and its associated fusing linear head, as shown in Figure~\ref{fig:fig2}. By combining the attention value from geometry priors and observation information, the model is equipped with both interaction awareness and visual awareness. 
For the human shape parameters $\beta$, we employ the same branch comprising 3 cross-attention layers. Moreover, we normalize the $\beta$ into $[-1,1]$ during processing.
Regarding the timesteps, following~\cite{stathopoulos2024score}, the input is conditioned on the timestep $t$ via scaling and shifting operations, defined as $\bm{x}_t = t_{s}\bm{x} + t_{b}$, where the parameters $t_s$ and $t_b$ are derived from an MLP encoder. 

With the help of both guidance $\bm{c}_{\rm I}$ and $\bm{c}_{\rm G}$, we can train the diffusion model with the objective:
\begin{equation}
    L_{\rm DM} = \mathbb{E}_{\bm{x}_{\rm 0}, \epsilon, t, \bm{c}_{\rm I}, \bm{c}_{\rm G}} || \epsilon - \epsilon_{\theta} (\bm{x}_t, t, \bm{c}_{\rm I}, \bm{c}_{\rm G} ) ||^{2}
    \label{eq: training}
\end{equation}

\subsection{Implementation}
Pytorch~\cite{paszke2019pytorch} is used for implementation. As for the image feature extractor, we employ the ResNet50~\cite{he2016deep} as a backbone. For the preliminary estimation of SMPL-H parameters, we adopt the Hand4Whole framework. Regarding the pre-trained affordance-aware network, we use PointNeXt~\cite{qian2022pointnext} and we freeze the model during training to maintain stability in object feature extraction. During the inference phase, DDIM sampling is performed for step size $\Delta t=2$, and the contact-driven iterative refinement is iterated for $N=10$ times. An intermediate noise level $\tau$ of 0.05 is selected. Further details regarding the hyper-parameters are provided in Section~\ref{sec: abla}.

\begin{figure*}[t]
    \centering
    \includegraphics[width=0.99\linewidth]{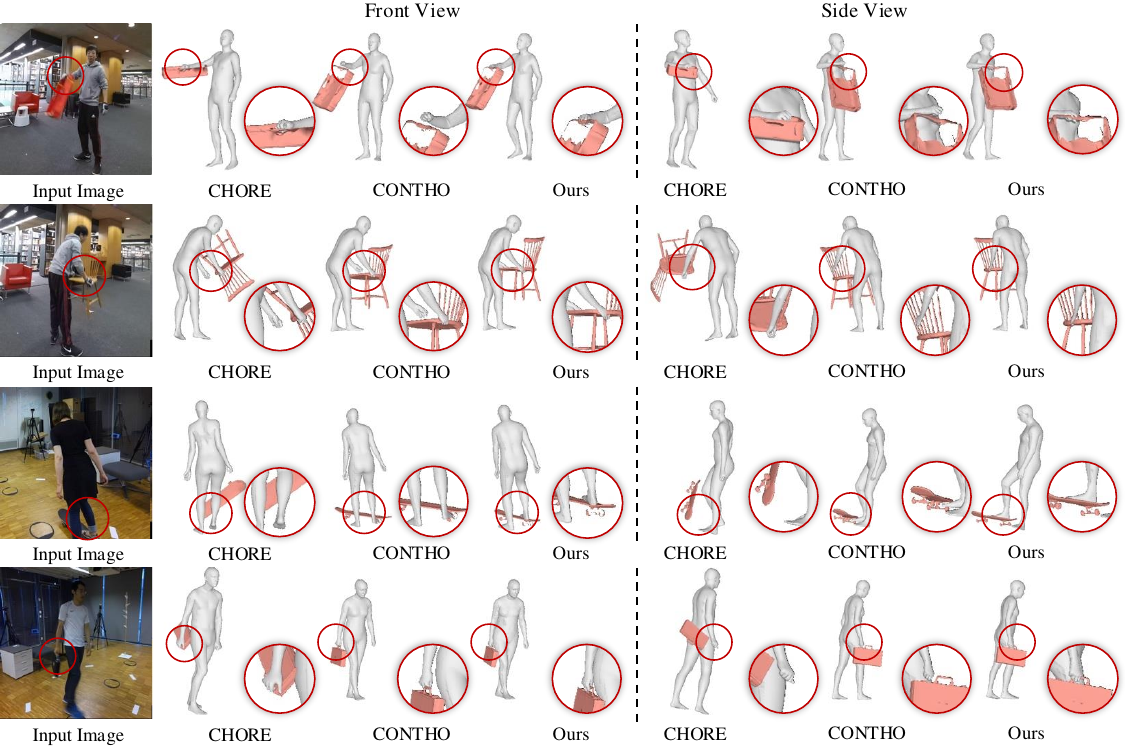}
    \vspace{-10pt}
    \caption{\textbf{Qualitative comparisons on BEHAVE~\cite{behave} and InterCap~\cite{intercap} benchmarks.} Our ScoreHOI recovers accurate human and object meshes with plausible interaction patterns. Notably, given only a single monocular image, the reconstruction results of the previous methods often appear unsatisfactory from the side view. By harnessing the inherent prior knowledge within generative models, our ScoreHOI effectively addresses this ill-posed problem, achieving superior reconstruction fidelity.}
    \label{fig:performance}
    \vspace{-15pt}
\end{figure*}
\section{Experiments}

To verify the effectiveness and efficiency of our proposed ScoreHOI, we conduct comprehensive experiments and ablation studies on the standard benchmarks. We will introduce the experimental settings and offer a comprehensive analysis through detailed comparisons and ablation studies.

\subsection{Experiment Setup}

\noindent \textbf{Datasets.}
\textbf{BEHAVE}~\cite{behave} dataset is the largest dataset of human-object interactions in natural environments, with 3D human, object and contact annotation. The dataset includes: (1) 8 subjects interacting with 20 objects in 5 natural environments. (2) In total 321 video sequences were recorded with 4 Kinect RGB-D cameras. (3) Textured scan reconstructions for the 20 objects. \textbf{InterCap}~\cite{intercap} contains 10 subjects (5 males and 5 females) interacting with 10 objects of various sizes and affordances. It includes contact with the hands or feet. In total, InterCap has 223 RGB-D videos, resulting in 67,357 multi-view frames, each containing 6 RGB-D images. \textbf{IMHD} dataset~\cite{imhd} contains 15 subjects (13 males, 2 females to participate in 10 different interaction scenarios. It also provides sequence-level textual guidance for each capture. Each split lasted from half a minute to one minute. It contains 32-view RGB videos with instance-level segmentation. IMHD utilizes ViTPose~\cite{xu2022vitpose} and MediaPipe to annotate 2D and 3D human key points. We \textit{only} utilize the IMHD training set during the diffusion training phase to augment the generation capability.
% Moreover, it provides 32-view RGB videos and instance-level segmentation, built on SAM~\cite{sem}, Track-Anything~\cite{track}, and XMem~\cite{xmem}.

\noindent \textbf{Training Details.} The training phase has two parts. Initially, we train the image backbone, the contact predictor and the vertices refinement at the first stage following the loss functions in ~\cite{nam2024joint}, including the supervision for the SMPL-H and object pose parameters, the contact mask and the 2D/3D vertices, etc. We train these baseline modules on BEHAVE~\cite{behave} and InterCap~\cite{intercap} training set. Subsequently, the diffusion model is trained using image features extracted from a frozen image backbone, incorporating IMHD~\cite{imhd} to enhance generative capabilities. Weight updates are facilitated by the Adam optimizer~\cite{kingma2014adam}, with a mini-batch size of 32 in the first stage and 256 in the second stage. In both stages, data augmentation techniques such as scaling, rotation, and color jittering are applied to enhance the diversity of the dataset. The training in the first stage is carried out for a total of 50 epochs, with an initial learning rate of $10^{-4}$, which is subsequently reduced by a factor of 10 after 30 epochs. In the second stage,  the parameters converge more rapidly, thereby necessitating only 30 epochs of training at a learning rate of $10^{-4}$. The model is trained using 4 NVIDIA RTX 4090 GPUs for around 1.5 days in total.

\noindent \textbf{Evaluation Details.} Following previous works~\cite{xie2022chore, zhang2020perceiving, nam2024joint}, we utilize two kinds of metrics to elevate the model's performance: chamfer distance, precision, and recall. \textbf{Chamfer distance~($\text{CD}_{\text{human}}$, $\text{CD}_{\text{object}}$).} Given the predicted 3D human and object meshes, we apply Procrustes alignment on combined 3D human and object meshes with the GT 3D human and object meshes. With the aligned 3D human and object meshes, we measure the Chamfer distance from GT separately on 3D human (CD$_{\text{human}}$) and 3D object (CD$_{\text{object}}$), in centimeters. \textbf{Precision recall and F-Score for contact from reconstruction ($\text{Contact}^{\text{rec}}_{\text{p}}$, $\text{Contact}^{\text{rec}}_{\text{r}}$, $\text{Contact}^{\text{rec}}_{\text{F-S}}$).} Precision measures how many of the predicted positive cases are positive. A high Precision means that when the model predicts a positive outcome, it is likely to be correct. Recall indicates how many of the actual positive cases were correctly identified by the model. A high Recall means that the model successfully identifies most of the positive cases. However, achieving high precision concurrently with high recall presents a significant challenge. Therefore we introduce $\text{F-Score}=2\times \text{p} \times \text{r} / (\text{p}+\text{r})$ as a metric to strike a balance between the two. We adopt these metrics for the reconstructed 3D human and object meshes to evaluate 3D human and object reconstruction, especially in terms of contact. We obtain a contact map by classifying human vertices within 5$cm$ of the object mesh following~\cite{nam2024joint}. Then, we measure the precision, recall and F-Score between the human contact map and the GT.

\begin{table}[t]
    \def\arraystretch{1.65}
    \renewcommand{\tabcolsep}{0.8mm}
    \caption{
    \textbf{Quantitative comparison of 3D human and object reconstruction with state-of-the-art methods on BEHAVE~\cite{behave} and InterCap~\cite{intercap}.} Our ScoreHOI model demonstrates superior accuracy, particularly in reconstructing contact interactions.
    }
    \small
    \vspace{-15pt}
    \begin{center}
    \scalebox{0.8}{
        \begin{tabular}{p{1.3cm}|>{\centering\arraybackslash}m{2.0cm}|>{\centering\arraybackslash}m{1.15cm}>{\centering\arraybackslash}m{1.15cm}>{\centering\arraybackslash}m{1.15cm}>{\centering\arraybackslash}m{1.15cm}>{\centering\arraybackslash}m{1.15cm}}
        \specialrule{.1em}{.05em}{0.0em}
        Datasets & \multicolumn{1}{c|}{Methods} & \multicolumn{1}{c}{\scalebox{0.9}{$\text{CD}_{\text{human}}{\downarrow}$}} & \multicolumn{1}{c}{\scalebox{0.9}{$\text{CD}_{\text{object}}{\downarrow}$}}  & 
        \multicolumn{1}{c}{\scalebox{0.8}{$\text{Contact}^{\text{rec}}_{\text{p}}{\uparrow}$}} & \multicolumn{1}{c}{\scalebox{0.8}{$\text{Contact}^{\text{rec}}_{\text{r}}{\uparrow}$}} & \multicolumn{1}{c}{\scalebox{0.8}{$\text{Contact}^{\text{rec}}_{\text{F-S}}{\uparrow}$}}\\ 
        \hline
        \multirow{4.25}{*}{BEHAVE} & PHOSA~\cite{zhang2020perceiving} & 12.17 & 26.62 & 0.393 & 0.266 & 0.317 \\
         & CHORE~\cite{xie2022chore} & 5.58 & 10.66 & 0.587 & 0.472 & 0.523\\
         & CONTHO~\cite{nam2024joint}  & 4.99 & 8.42 & 0.628 & 0.496 & 0.554\\ 
        &\cellcolor[gray]{0.9} Ours  &\cellcolor[gray]{0.9} \textbf{4.85} &\cellcolor[gray]{0.9} \textbf{7.86} &\cellcolor[gray]{0.9} \textbf{0.634} &\cellcolor[gray]{0.9} \textbf{0.586} &\cellcolor[gray]{0.9} \textbf{0.609}\\ 
         \hline
         \multirow{4.25}{*}{InterCap} & PHOSA~\cite{zhang2020perceiving} & 11.20 & 20.57 & 0.228 & 0.159 & 0.187 \\ 
         & CHORE~\cite{xie2022chore} & 7.01 & 12.81 & 0.339 & 0.253 & 0.290\\
         & CONTHO~\cite{nam2024joint}  & 5.96 & 9.50 & \textbf{0.661}  & 0.432 & 0.522\\
         &\cellcolor[gray]{0.9} Ours &\cellcolor[gray]{0.9} \textbf{5.56} &\cellcolor[gray]{0.9} \textbf{8.75} &\cellcolor[gray]{0.9} 0.627  &\cellcolor[gray]{0.9} \textbf{0.590} &\cellcolor[gray]{0.9} \textbf{0.578}\\
        \specialrule{.1em}{-0.05em}{-0.05em}
        \end{tabular}
    }
    \end{center}

    \vspace{-10pt}
    \label{tab:sota}
\end{table}

\begin{table}[t]
    \def\arraystretch{1.2}
    \renewcommand{\tabcolsep}{0.8mm}
    \caption{
    \textbf{Quantitative comparison of optimization efficiency.} While achieving superior performance, our ScoreHOI maintains a higher inference efficiency compared with previous methods.}
    \vspace*{-15pt}
    \small
    \begin{center}
        \begin{tabular}
            {>{\centering\arraybackslash}m{2.4cm}|>{\centering\arraybackslash}m{1.35cm}>{\centering\arraybackslash}m{1.35cm}>{\centering\arraybackslash}m{1.35cm}}
            \specialrule{.1em}{.05em}{0.0em}
            Methods & \scalebox{1.0}{$\text{CD}_{\text{human}}{\downarrow}$} & \scalebox{1.0}{$\text{CD}_{\text{object}}{\downarrow}$}  & 
            \scalebox{1.0}{$\text{FPS}{\uparrow}$} \\ 
            \hline
            CHORE~\cite{xie2022chore} & 5.58 & 10.66 & 0.0035 \\
            VisTracker~\cite{vistracker} & 5.24 & 7.89 & 0.0359 \\
            \rowcolor[gray]{0.9}
            Ours-Faster & 4.87 & 7.95 & \textbf{2.0080} \\
            \rowcolor[gray]{0.9}
            Ours & \textbf{4.85} & \textbf{7.86} & 0.2895 \\
            \specialrule{.1em}{-0.05em}{-0.05em}
        \end{tabular}
    \end{center}
    \vspace*{-15pt}
    \label{tab: efficiency}
\end{table}

\subsection{Comparison with State-of-the-art Methods}

\noindent \textbf{Performance Comparison.} We compare ours with previous state-of-the-art methods with two experimental protocols, BEHAVE~\cite{behave} and InterCap~\cite{intercap}. Quantitative results are recorded in Table~\ref{tab:sota}. Our method achieves the best performance in most metrics. To harmonize contact precision and recall, where the objective is to maximize both values, we introduce the contact F-Score as a metric to assess the quality of contact reconstruction. Due to the comprehensive physical constraints, we surpass the previous state-of-the-art method by 9\% on contact F-score. Additionally, the optimization results are visualized in \Cref{fig:performance}, illustrating that our ScoreHOI reconstructs contact relationships with greater rationality and physical plausibility. It is noteworthy that reconstruction inaccuracies often arise in side views given only a single monocular image. Our ScoreHOI effectively mitigates this ill-posed problem by harnessing the prior knowledge within generative models.

\begin{table}[t]
    \def\arraystretch{1.42}
    \renewcommand{\tabcolsep}{0.8mm}
    \caption{
    \textbf{Ablation studies of modules, conditions and guidance.} The CDIR refers to the contact-driven iterative refinement. All of the modules, conditions and guidance strategies we design contribute to the improvement of our model.
    }
    \vspace{-15pt}
    \small
    \begin{center}
    \resizebox{\columnwidth}{!}{
        \begin{tabular}
            {>{\raggedright\arraybackslash}m{2.0cm}|>{\centering\arraybackslash}m{1.2cm}>{\centering\arraybackslash}m{1.2cm}>{\centering\arraybackslash}m{1.2cm}>{\centering\arraybackslash}m{1.2cm}>{\centering\arraybackslash}m{1.2cm}}
            \specialrule{.1em}{.05em}{0.0em}
            Methods & \scalebox{0.9}{$\text{CD}_{\text{human}}{\downarrow}$} & \scalebox{0.9}{$\text{CD}_{\text{object}}{\downarrow}$}  & 
            \scalebox{0.85}{$\text{Contact}^{\text{rec}}_{\text{p}}{\uparrow}$} & \scalebox{0.85}{$\text{Contact}^{\text{rec}}_{\text{r}}{\uparrow}$} & \scalebox{0.85}{$\text{Contact}^{\text{rec}}_{\text{F-S}}{\uparrow}$} \\ 
            \hline
            \textit{* Module} \\
            w/o diffusion & 5.03 & 8.48 & 0.612 & 0.523 & 0.588 \\
            w/o CDIR & 4.93 & 7.98 & 0.628 & 0.545 & 0.577 \\
            \hline
            \textit{* Condition} \\
            No condition & 4.94 & 8.23 & 0.626 & 0.549 & 0.585 \\
            w/o $\bm{c}_{\rm G}$ & 4.87 & 7.99 & 0.628 & 0.559 & 0.591\\
            w/o $\bm{c}_{\rm I}$ & 4.88 & 8.03 & 0.631 & 0.566 & 0.597\\
            \hline
            \textit{* Guidance} \\
            No guidance & 4.93 & 8.01 & 0.624 & 0.524 & 0.570 \\
            w/o $L_{ho}$ & 4.87 & 7.95 & 0.632 & 0.525 & 0.574 \\
            w/o $L_{pt}$ & 4.87 & 7.93 & 0.619 & 0.567 & 0.592 \\
            w/o $L_{of}$ & 4.89 & 7.95 & 0.631 & 0.577 & 0.602 \\
            \rowcolor[gray]{0.9}
            Full model  & \textbf{4.85} & \textbf{7.86} & \textbf{0.634} & \textbf{0.586} & \textbf{0.609} \\ 
            \specialrule{.1em}{-0.05em}{-0.05em}
        \end{tabular}
    }
    \end{center}
    \vspace{-20pt}
    \label{tab:abla}
\end{table}

\noindent \textbf{Efficiency Comparison.} To evaluate the inference efficiency of various optimization methods, we measure the Frames Per Second (FPS) on a single Nvidia RTX 4090 GPU. Our ScoreHOI is benchmarked against two established methods, CHORE~\cite{xie2022chore} and VisTracker~\cite{vistracker}, both of which utilize the Adam Optimizer for reconstructing image and video inputs, respectively. The evaluation results, as shown in Table~\ref{tab: efficiency}, indicate that our method’s efficiency surpasses that of the comparator methods by one to two orders of magnitude while simultaneously delivering superior performance. Furthermore, we assess a more expedient framework with $N=2$ as detailed in the 4-th row, Table~\ref{tab:abla_tnt}. This configuration achieves a substantial seven-fold increase in efficiency with minimal compromise in performance.

\begin{figure*}
    \centering
    \includegraphics[width=0.99\linewidth]{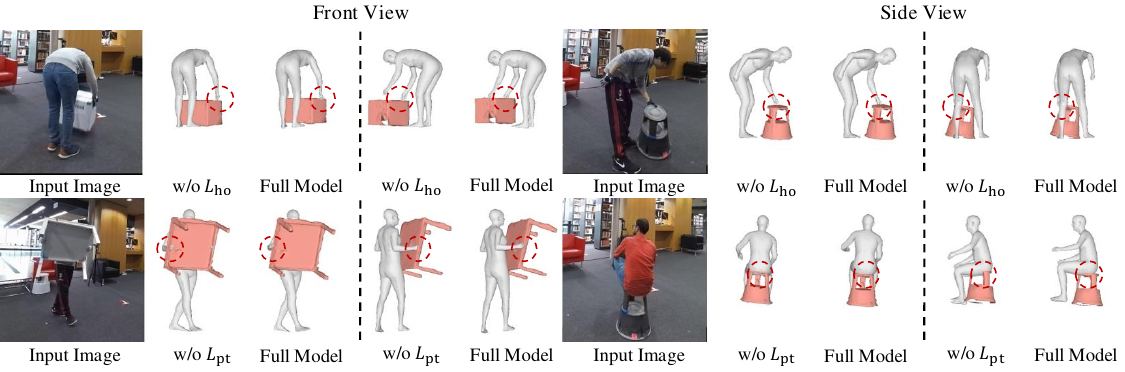}
    \vspace{-5pt}
    \caption{\textbf{Qualitative results for ablation study.} Upper row: the ablation study of $L_{\rm ho}$. The inclusion of contact guidance between the human and object significantly enhances the physical plausibility of the interactions. Bottom row: the ablation study of $L_{\rm ho}$. The absence of a penetration penalty results in a notable rise in the occurrence of unreasonable interactions.}
    \vspace{-10pt}
    \label{fig: abla}
\end{figure*}
\subsection{Ablation Study}
\label{sec: abla}
We conduct all ablation studies on the BEHAVE~\cite{behave} test split as a standard benchmark for fair comparison.

\noindent \textbf{Effectiveness of Optimizer Modules.} We evaluate the influence of the diffusion sampling loop and the proposed contact-driven iterative refinement, as shown in the upper part of Table~\ref{tab:abla}. Utilizing an affordance-aware regressor, our backbone network is capable of achieving satisfactory outcomes independently of the diffusion model. Nonetheless, there remains potential for enhancement, particularly in terms of contact recall. Additionally, we perform an ablation study excluding CDIR, wherein the contact map is not updated during the sampling loops. The results indicate that CDIR contributes to the improvement of both reconstruction accuracy and contact quality.

% \begin{table}[t]
%     \def\arraystretch{1.0}
%     \caption{Experimental Results with Three Variables and Three }
%     \label{tab:abla_tnt}
%     \vspace{-5pt}
%     \begin{center}
%     \scalebox{1.0}{
%         \begin{tabular}{ccc|ccc}
%         \toprule
%         $N$ & $\tau$ & $\Delta t$ & $\text{CD}_{\text{human}}{\downarrow}$& $\text{CD}_{\text{object}}{\downarrow}$ & $\text{Contact}^{\text{rec}}_{\text{F-S}}{\uparrow}$ \\
%         \hline
%         20 & 50 & 25 & - & - & - \\
%         20 & 50 & 10 & - & - & - \\
%         20 & 50 & 5 & - & - & - \\
%         \hline
%         2 & 50 & 2 & - & - & - \\
%         5 & 50 & 2 & - & - & - \\
%         10 & 50 & 2 & - & - & - \\
%         \hline
%         20 & 25 & 2 & - & - & - \\
%         20 & 100 & 2 & - & - & - \\
%         20 & 50 & 2 & - & - & - \\
%         \bottomrule
%         \end{tabular}
%     }
%     \end{center}
% \end{table}

\begin{table}[t]
    \def\arraystretch{1.2}
    \renewcommand{\tabcolsep}{0.8mm}
    \caption{
    \textbf{Ablation studies of optimization hyper-parameters.} We evaluate the performance across various optimization hyper-parameter configurations. The optimal results measured by chamfer-distance are selected to establish our baseline.}
    \vspace*{-15pt}
    \small
    \begin{center}
    % \resizebox{\columnwidth}{!}{
        \begin{tabular}
            {>{\centering\arraybackslash}m{0.75cm}>{\centering\arraybackslash}m{0.75cm}>{\centering\arraybackslash}m{0.75cm}|>{\centering\arraybackslash}m{1.35cm}>{\centering\arraybackslash}m{1.35cm}>{\centering\arraybackslash}m{1.35cm}}
            \specialrule{.1em}{.05em}{0.0em}
            $N$ & $\tau$ & $\Delta t$ & \scalebox{0.9}{$\text{CD}_{\text{human}}{\downarrow}$} & \scalebox{0.9}{$\text{CD}_{\text{object}}{\downarrow}$}  & 
            \scalebox{0.9}{$\text{Contact}^{\text{rec}}_{\text{F-S}}{\uparrow}$} \\ 
            \hline
            10 & 50 & 25 & 5.80 & 10.70 & 0.539 \\
            10 & 50 & 10 & 5.22 & 8.72 & 0.584 \\
            10 & 50 & 5 & 4.99 & 8.16 & 0.569 \\
            \hline
            2 & 50 & 2 & 4.87 & 7.95 & 0.604 \\
            5 & 50 & 2 & 4.86 & 7.87 & 0.607 \\
            20 & 50 & 2 & 4.87 & 7.89 & 0.610 \\
            \hline
            10 & 25 & 2 & 4.87 & 7.95 & 0.606 \\
            10 & 100 & 2 & 4.88 & 8.07 & \textbf{0.615} \\
            \rowcolor[gray]{0.9}
            10 & 50 & 2 & \textbf{4.85} & \textbf{7.86} & 0.609 \\
            \specialrule{.1em}{-0.05em}{-0.05em}
        \end{tabular}
    % }
    \end{center}
    \vspace*{-15pt}
    \label{tab:abla_tnt}
\end{table}

\noindent \textbf{Effectiveness of Conditions.} As illustrated in the central section of Table~\ref{tab:abla}, the exclusion of either image guidance or geometry guidance from the diffusion model leads to a decrement in reconstruction performance. In comparison to the setting in the first row, which excludes the diffusion module, our contact-driven iterative refinement approach demonstrates the capability to enhance performance even in the absence of additional conditions.

\noindent \textbf{Effectiveness of Physical Guidance.} We investigate the impact of varying physical constraints during DDIM sampling, as illustrated at the bottom of Table~\ref{tab:abla}. Experimental findings indicate that the incorporation of three distinct objectives enhances both reconstruction accuracy and contact performance. The omission of $L_{\rm ho}$ results in a substantial decline in recall of the contact map, thereby diminishing the occurrence of contact interactions in the outputs. Contrarily, the absence of $L_{\rm pt}$, leads to a reduction in contact precision, attributable to an increase in penetrations and the redundant contact of unrelated parts. To substantiate our findings, qualitative results are also presented in Figure~\ref{fig: abla}.

\noindent \textbf{Analysis of Optimization Hyper-Parameters.} We undertake an ablation analysis to investigate the influence of optimization hyper-parameters, including the iteration times $N$, the intermediate noise level $\tau$, and the DDIM step size $\Delta t$, as depicted in Table~\ref{tab:abla_tnt}.  Our findings indicate that an increase in $N$ results in a higher contact F-Score, attributable to the greater number of refinement steps enhancing the contact fidelity. A similar trend is observed with $\tau$, where elevated values correspond to improved contact relationships. Regarding $\Delta t$, more sampling steps is observed to produce substantially superior results. In summary, our results reveal a balance between reconstruction quality and interaction fidelity, and we finally choose $N=10, \tau=50, \Delta t = 2$ as the optimal baseline parameter combination.

\section{Conclusion and Future Work}

In this paper, we have introduced ScoreHOI, an innovative and effective framework tailored for the physically plausible reconstruction of human-object interactions. It integrates diffusion-based prior knowledge for score-guided sampling refinement and incorporates physical constraints for realism. We also propose a contact-driven iterative refinement algorithm to improve the recovery of contact patterns. Extensive evaluations show superior reconstruction accuracy, enhanced physical plausibility, and faster inference than optimization-based methods. Nevertheless, we acknowledge our model has limitations in generalization due to the pre-defined canonical pose of known objects in the training dataset, which prevents the optimization of parameters for objects with undefined canonical pose. Our future research will focus on solving the issue of template-free unseen objects. We anticipate that our work will contribute a novel perspective to the field of joint reconstruction of human-object interactions, fostering further research and development in this area.

\section{Acknowledgement}
This work was supported by Guangdong Natural Science Funds for Distinguished Young Scholar~(No. 2025B1515020012) and Shenzhen Science and Technology Program~(JCYJ20240813111903006).
% an innovative and effective framework tailored for the physically plausible reconstruction of human-object interactions. By incorporating diffusion-based prior knowledge of human-object interactions, we facilitate the refinement of coarse estimation results through score-guided sampling. To ensure the sampling process adheres to physical realism, we integrate physical constraints into the framework. Additionally, we propose a contact-driven iterative refinement algorithm aimed at enhancing the recovery of contact interaction patterns. Extensive quantitative and qualitative experimental evaluations demonstrate that our framework achieves superior accuracy in reconstruction and greater physical plausibility in contact interaction. Furthermore, our ScoreHOI exhibits a significantly faster inference speed compared to optimization-based algorithms. We anticipate that our work will contribute a novel perspective to the field of joint reconstruction of human-object interactions, fostering further research and development in this area.
%-------------------------------------------------------------------------

{\small
\bibliographystyle{ieee_fullname}
\bibliography{main}
}

\maketitlesupplementary

% \afterpage{\clearpage}
\setcounter{page}{1}

\setcounter{section}{0}
\renewcommand\thesection{\Alph{section}}
\setcounter{figure}{0}
\renewcommand\thefigure{\Alph{figure}}
\setcounter{table}{0}
\renewcommand\thetable{\Alph{table}}
\setcounter{algorithm}{0}
\renewcommand\thealgorithm{\Alph{algorithm}}

In this appendix, we describe the DDIM sampling loop in our methods in pseudo-code in Section~\ref{sec: ddim}. We also provide additional detailed implementations and qualitative comparisons in Section~\ref{sec: detailed} and Section~\ref{sec: more_qual}. Furthermore, we append rendered 3D object models and the source code along with this appendix.

\section{DDIM Refinement Loop}
\label{sec: ddim}
For a deeper understanding of the DDIM sampling loop, we illustrate a pseudo-code implementation in Algorithm~\ref{algo:sgddim}. The algorithm describe the $\text{DDIM\_Loop}$ in detail.

\section{Detailed Implementations}
\label{sec: detailed}
We employ the PointNeXt~\cite{qian2022pointnext} model, pre-trained on the ModelNet40~\cite{wu20153d} dataset, as our affordance-aware regressor. ModelNet40 is a extensively utilized benchmark dataset for 3D shape classification and retrieval tasks, comprising 12,311 CAD models across 40 distinct object categories, including airplanes, chairs, tables, and cars. PointNeXt maintains the fundamental hierarchical architecture of PointNet++~\cite{qi2017pointnet++} while integrating contemporary deep learning techniques to augment performance and efficiency. Through the incorporation of point cloud awareness, the model is capable of comprehending object geometry across a variety of human-object interaction scenarios.
For the contact predictor and mesh regressor, we leverage the contact estimation transformer and the contact-based refinement methodology proposed by~\cite{nam2024joint}. The contact predictor receives human and object feature tokens, along with estimated human and object mesh vertices, as inputs to generate contact masks through two symmetrical 4-layer transformer blocks. During our contact-driven iterative refinement process, we iteratively update the human and object meshes to enhance contact interactions, thereby refining the contact prediction results with each iteration. After $N$ iterations of optimization, we obtain the $\bm{x}^N_0$ for final refinement. The mesh regressor, also constructed with two analogous 4-layer transformer branches, takes the updated human and object feature tokens and contact masks as input to further refine the mesh results.

\section{Ablation Studies}
\begin{figure}[t]
    \centering
    \includegraphics[width=0.99\linewidth]{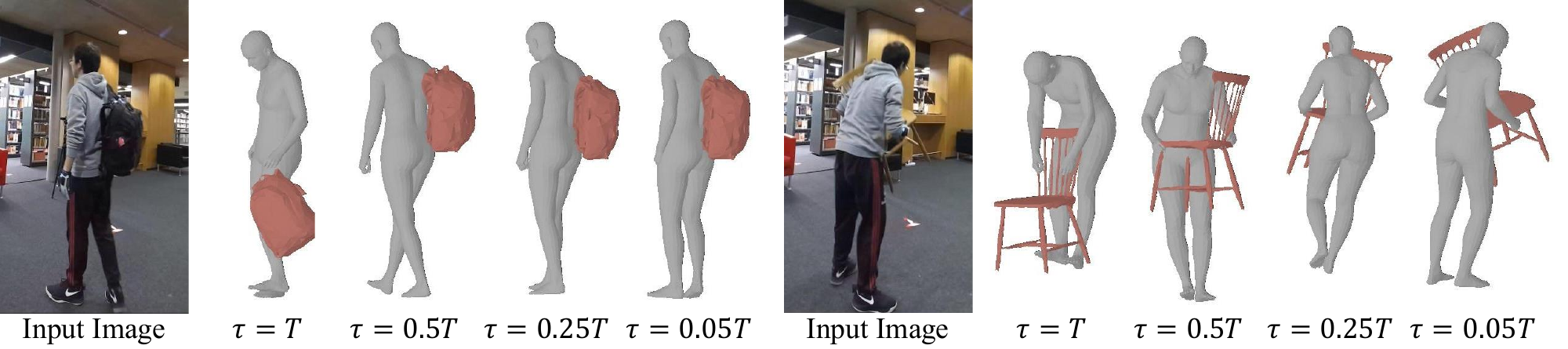} 
    \vspace{-10pt}
    \caption{Generation results starting from different noise levels.}
    \label{fig: reb_generation}
    \vspace{-10pt}
\end{figure}

\noindent
\textbf{About Diffusion Priors}. We demonstrate the generation results starting from different noise levels without physical guidance, as shown in the Figure~\ref{fig: reb_generation}. The IG-Adapter successfully constructs plausible human-object interaction patterns, confirming that the diffusion model has acquired a valid prior distribution.

% \begin{wraptable}{r}{5.5cm}
% \vspace{-10pt} 
% % \centering
% \scriptsize
% \renewcommand\arraystretch{0.65}
% \setlength{\tabcolsep}{2.pt}
% \hspace{-.2in}
%  \adjustbox{width=1.1\linewidth}
%     { 
%         \begin{tabular}
%             {>{\raggedright\arraybackslash}m{0.6cm}|>{\centering\arraybackslash}m{1.0cm}>{\centering\arraybackslash}m{1.0cm}>{\centering\arraybackslash}m{1.0cm}>{\centering\arraybackslash}m{1.0cm}}
%             \specialrule{.1em}{.05em}{0.0em}
%              & \scalebox{0.8}{HDM} & \scalebox{0.8}{InterTrack} & \scalebox{0.8}{ScoreHOI} & \scalebox{0.8}{ScoreHOI-F} \\ 
%             \hline
%             $\text{FPS}{\uparrow}$  & 0.0047 & 0.0012 & 0.2895 & \textbf{2.0080} \\ 
%             \specialrule{.1em}{-0.05em}{-0.05em}
%         \end{tabular}
%     }
%     \label{tab:reb_templatefree}
% \vspace{-15pt} 
% \end{wraptable}

\begin{table}[t]
\scriptsize
\centering
\renewcommand\arraystretch{0.65}
\setlength{\tabcolsep}{2.pt}
\caption{Efficiency comparision with template-free methods.}
\vspace{-10pt}
 \adjustbox{width=0.99\linewidth}
    { 
        \begin{tabular}
            {>{\raggedright\arraybackslash}m{0.6cm}|>{\centering\arraybackslash}m{1.0cm}>{\centering\arraybackslash}m{1.0cm}>{\centering\arraybackslash}m{1.0cm}>{\centering\arraybackslash}m{1.0cm}}
            \specialrule{.1em}{.05em}{0.0em}
             & \scalebox{0.75}{HDM} & \scalebox{0.75}{InterTrack} & \scalebox{0.75}{ScoreHOI} & \scalebox{0.75}{ScoreHOI-F} \\ 
            \hline
            $\text{FPS}{\uparrow}$  & 0.0047 & 0.0012 & 0.2895 & \textbf{2.0080} \\ 
            \specialrule{.1em}{-0.05em}{-0.05em}
        \end{tabular}
    }
    \label{tab:reb_templatefree}
\end{table}

\noindent
\textbf{Comparison with template-free methods.} 
We compared ScoreHOI with template-free methods, including HDM~\cite{xie2024template} and InterTrack~\cite{xie2024InterTrack}. As shown in the Table~\ref{tab:reb_templatefree}, ScoreHOI surpasses these methods in efficiency. Additionally, these methods output point clouds, which are less practical for applications like robotic data collection due to the additional time and uncertainty introduced when fitting SMPL parameters.

\section{More Qualitative Results}
\label{sec: more_qual}

\begin{figure*}
    \centering
    \includegraphics[width=0.99\linewidth]{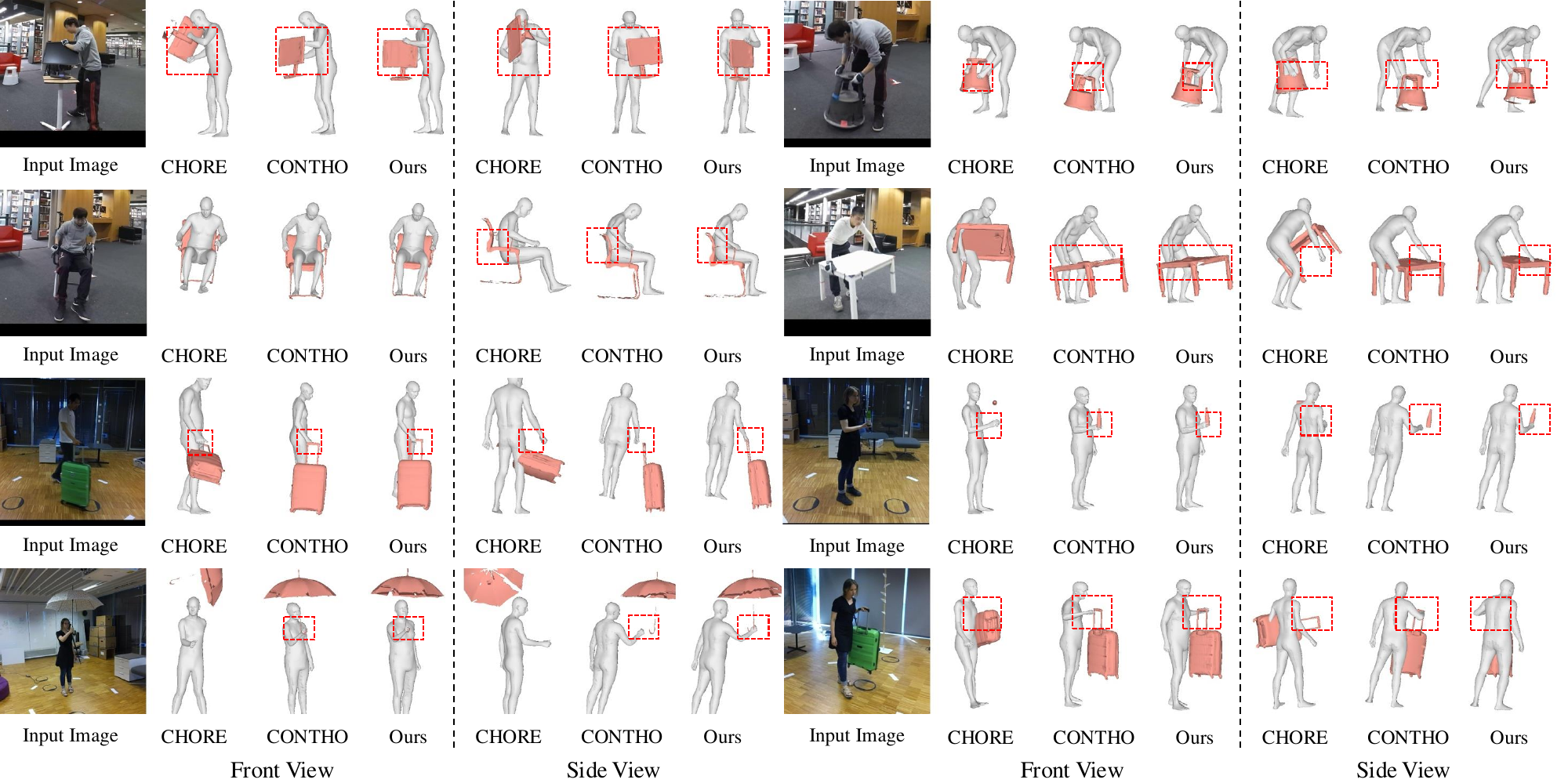}
    % \vspace{-10pt}
    \caption{\textbf{Extra Qualitative comparisons.} We highlight the contact interaction within each picture.}
    \label{fig:performance_supp}
    % \vspace{-15pt}
\end{figure*}

To further substantiate the performance of our ScoreHOI, we present additional qualitative results of human-object interaction reconstruction in Figure~\ref{fig:performance_supp}. Our methodology exhibits superior performance across diverse interaction patterns, such as sitting, carrying, grasping, and lifting. By incorporating physical constraints, our ScoreHOI achieves a higher degree of accuracy and physical fidelity in the reconstruction of human and object meshes.

\section{Geometry Model Demos}

The demo 3D geometry mesh results are available under the \textbf{demo} directory. Reviewers are able to assess the performance from any perspective utilizing software such as MeshLab or Blender.

\begin{algorithm*}[!htbp]
	\caption{Score-Guided DDIM refinement loop}
     \label{algo:sgddim}
	\begin{algorithmic}[1]
		\State \textbf{Input:} latent parameters $\bm{x}^n_0$ at step $n$, denoising model $\epsilon_{\phi}$, image features $\bm{c}_{\rm I}$, geometry features $\bm{c}_{\rm G}$, gradient step size $\rho$, noise level $\tau$, DDIM step size $\Delta t$, estimated caontact masks $\{\mathbf{M}_{i}\}_{i \in \{ {\rm h}, {\rm o} ,{\rm f}\}}$
            \State \textbf{Output:} latent parameters $\bm{x}^{n+1}_0$ for next sampling step $n+1$
            \State $\bm{x}_{\tau}=\text{DDIMInvert}(\bm{x}^n_{0}, \bm{c}_{\rm I}, \bm{c}_{\rm G})$ \Comment{ Run DDIM inversion until noise level $\tau$}
            \For{$t=\tau$ to $\Delta t$ with step size $\Delta t$}
                \State $\tilde\epsilon \gets \epsilon_{\phi}(\bm{x}^n_t, t, \bm{c}_{\rm I}, \bm{c}_{\rm G})$ \Comment{ Predict noise}
                \State Initialize computational graph for $\bm{x}^n_t$
                \State $\hat{\bm{x}^n}_{0} \gets \frac{1}{\sqrt{\alpha_{t}}} (\bm{x}^n_t - \sqrt{1 - \alpha_{t}}) \tilde\epsilon$ \Comment{ Predict one-step denoised result}
                \State $L_{\mathcal{P}} \gets \text{PhysicalGuidance}(\hat{\bm{x}}^n_{0},\{\mathbf{M}_{i}\}_{i \in \{ {\rm h}, {\rm o} ,{\rm f}\}})$ \Comment{ Compute physical guidance loss}
                \State $\tilde\epsilon^{'} \gets \tilde\epsilon + \rho\sqrt{1 - \alpha_{t}}\nabla_{\bm{x}^n_t}L_{\mathcal{P}}$ \Comment{ Compute modified noise after score-guidance}
                \State $\hat{\bm{x}^n}_{0}^{'} \gets \frac{1}{\sqrt{\alpha_{t}}} (\bm{x}^n_t - \sqrt{1 - \alpha_{t}}) \tilde\epsilon^{'}$ \Comment{ Predict one-step denoised result with modified noise}
                \State $\bm{x}^n_{t-\Delta t} \gets {\sqrt{\alpha_{t-\Delta t}}}\hat{\bm{x}}_{0}^{n'} + \sqrt{1 - \alpha_{t-\Delta t}} \tilde\epsilon^{'}$ \Comment{ DDIM sampling step}
            \EndFor
            \State $\bm{x}^{n+1}_{0} \gets \hat{\bm{x}}_{0}^{n'}$ \Comment{Update $\bm{x}^{n}_{0}$ for next generation}
		\State \Return $\bm{x}^{n+1}_{0}$
	\end{algorithmic}  
\end{algorithm*}

% \begin{algorithm}[t]
%     \caption{Contact-Driven Iterative Refinement}
%     \label{algo: contact}
%     \begin{algorithmic}[1]
%     \State \textbf{Input:} Initial parameters $\bm{x}^0_0$, diffusion model $\epsilon_{\theta}$, image feature $\mathcal{F}$, time steps $t$, and condition $\bm{c}$
%     \State \textbf{Output:} The optimized parameters $\bm{x}^N_0$ 
%     \For{$n = 0$ to $N-1$} 
%         \State $\mathcal{F}^{n}_{\rm h}, \mathcal{F}^{n}_{\rm o} \gets \textbf{sample}(\bm{x}^n_0, \mathcal{F})$
%         \State $\{\mathbf{M}_{i}\}_{i \in \{ {\rm h}, {\rm o} ,{\rm f}\}} \gets \textbf{Contact}(\bm{x}^n_0, \mathcal{F}^n_{\rm h}, \mathcal{F}^n_{\rm o})$ 
%         \State $\bm{x}^{n+1}_0 \gets \textbf{DDIM\_Loop}(\bm{x}^n_0, t, \bm{c}, \epsilon_{\theta}, \{\mathbf{M}_{i}\}_{i \in \{ {\rm h}, {\rm o} ,{\rm f}\} })$
%     \EndFor
%     \State \Return $\bm{x}^N_0$
%     \end{algorithmic}
% \end{algorithm}

\end{document}